\documentclass{article}
\usepackage{amsmath}
\usepackage{graphicx} 
\usepackage{blindtext} 
\usepackage{balance}
\usepackage[sc]{mathpazo} 
\usepackage[T1]{fontenc}  
\usepackage{microtype}    
\usepackage{adjustbox}
\usepackage[english]{babel} 

\usepackage[hmarginratio=1:1,top=32mm,columnsep=20pt]{geometry} 
\usepackage[hang, small,labelfont=bf,up,textfont=it,up]{caption} 
\usepackage{booktabs} 

\usepackage{lettrine} 

\usepackage{enumitem} 
\setlist[itemize]{noitemsep} 

\usepackage{abstract} 

\usepackage{titlesec} 
\renewcommand\thesection{\Roman{section}} 
\renewcommand\thesubsection{\roman{subsection}} 
\titleformat{\section}[block]{\large\scshape\centering}{\thesection.}{1em}{} 
\titleformat{\subsection}[block]{\large}{\thesubsection.}{1em}{} 

\usepackage{fancyhdr} 
\usepackage{titling} 

\usepackage{hyperref} 

\pretitle{\begin{center}\Huge\bfseries} 
\posttitle{\end{center}} 
\title{Comparison Study Between Token Classification and Sequence Classification In Text Classification} 
\author{%
\textsc{Amir Jafari} \\[1ex] 
}
\date{\today} 

\usepackage{float}
\usepackage{array}
\newcolumntype{M}[1]{>{\centering\arraybackslash}m{#1}}
\usepackage{multirow}
\newcolumntype{C}[1]{>{\centering\arraybackslash}p{#1}}
\usepackage[justification=centering]{caption}

\renewcommand{\maketitlehookd}{%

\begin{abstract}
\noindent Unsupervised Machine Learning techniques have been applied to Natural Language Processing tasks and surpasses the benchmarks such as GLUE with great success. Building language models approach achieves good results in one language and it can be applied to multiple NLP task such as classification, summarization, generation and etc as an out of box model. Among all the of the classical approaches used in NLP, the masked language modeling is the most used. In general, the only requirement to build a language model is presence of the large corpus of textual data. Text classification engines uses a variety of models from classical and state of art transformer models to classify texts for in order to save costs. Sequence Classifiers are mostly used in the domain of text classification. However Token classifiers also are viable candidate models as well. Sequence Classifiers and Token Classifier both tend to improve the classification predictions due to the capturing the context information differently. This work aims to compare the performance of Sequence Classifier and Token Classifiers and evaluate each model on the same set of data. In this work, we are using a pre-trained model as the base model and Token Classifier and Sequence Classier heads results of these two scoring paradigms with be compared..
\end{abstract}
}

\begin{document}

\maketitle
\section{Introduction}

Text classification engines has been used as an alternative and complement to human annotation for many years. Essay tests are an example of a text responses where students are given a particular prompt topic to write about and elaborate on. These texts are evaluated for their writing quality in areas such as grammar, fluency of elaboration and how well they are organized. However, evaluating texts in large amount is impractical and classifying it is mostly very complicated and difficult \cite{attali2006automated}.

Text classifications engines can't genuinely read and comprehend texts the way humans or raters can. Text classification engines do not use direct qualities of a text such as how fluent it is or how grammatically correct. But human raters may explicitly evaluate numerous of those intrinsic variables to produce a prediction. This gives a clear distinction between machine scoring and human labeling. In \cite{goh2020evaluating} they studied whether human scoring or machine scoring classification models are better at classifying scientific research abstracts according to a fixed set of discipline groups. Text engines are mostly trained by collecting generally a few couple of thousand sample text which were previously assessed by human raters. There has always been skepticism over machine scoring over time mostly in regard to the fact that computer models does not understand the written text as they human does \cite{page1995computer}. However, by looking at labeling between human raters and machine models and create a an statical metric like Cohen’s Quadratic Weighted Kappa score we could measure the quality of scoring models (human vs machine) as long as we prove the sample data is statistically rules significant.

There are two major categories of text classification engines available in the field natural language processing: traditional and neural network-based models. Traditional text classification engines examine a text using carefully crafted features. The effectiveness of such models is closely correlated with the caliber of the underlying components. The most important element of such engines is manually designing the features and rules, which is time-consuming. A feature free method for learning the relationship between a text assigned label in \cite{taghipour2016neural} investigates  the use of recurrent neural network model as a modern methods. For the purpose of text classification engine, various neural network models are used in the field \cite{liang2018automated, rodriguez2019language, uto2020robust, farag2018neural}.

In comparison to statistical models with handmade characteristics, neural network models have recently been utilized for text classifiers \cite{alikaniotis2016automatic, dong2016automatic, taghipour2016neural} have produced better results. Word representations in particular are utilized as the input to a neural network model which produces a single dense vector for each word representation in the whole text. Based on a non-linear transformation in each layer on the representation, a label is given. It has been demonstrated that across several domains, neural network models are more reliable than statistical models without the use of handmade features.

Text classifiers engines have been using both convolutional neural networks \cite{chen2015convolutional} and recurrent neural networks \cite{mikolov2010recurrent} modles. A single-layer LSTM also has been used over the word sequence to model the text, and they separately model sentences and documents using a two-level hierarchical CNN structure. It is well known that LSTMs excel at modeling temporal and ordered data while CNNs excel at capturing local information. The effectiveness of LSTMs and CNNs using the identical text classification settings has been discussed \cite{dong2016automatic}.

Pre-trained language models have recently been used to significantly increase the performance of phrase prediction tasks by combining representations from many layers, creating an auxiliary sentence, using multitask learning, etc.

Large pre-trained language models have recently demonstrated amazing abilities in representation and generalization, like GPT \cite{radford2018improving}, BERT \cite{devlin2018bert}, XLNet \cite{dong2017attention}, etc. These models now perform a variety of downstream tasks including text summarization, generation, classification, and regression with enhanced accuracy. There are numerous innovative methods for improving already-trained language models. Auxiliary sentence construction was suggested by \cite{sun2019fine} to improve sentiment classification problems. To address sequential sentence classification tasks, Cohan \cite{cohan2019pretrained} introduced additional distinct tokens to obtain representations of each sentence. A work in \cite{sun2019fine} on several fine-tuning techniques, such as combining text representations from various layers and leveraging multi-task learning has been done.

It makes sense to use BERT type models or generally transformer-based models to learn text representations given the enormous success of pre-trained language models like BERT in learning text representations with deep semantics. The BERT approach, which relies heavily on self-attention, can capture interactions between any two words throughout the entire body of articles (long texts). Previous work on \cite{sun2019fine}  demonstrates that merging text representations from various layers does not significantly increase performance. The length of texts for the text classification job is close to the maximum allowed by the BERT model, making it challenging to create an auxiliary phrase.

In this article, we suggest using two type classification heads for AES engine to our base model to learn text representations that fully reflect deep semantics.  These two models are Token Classifier and Sequence Classier models. A token classifier uses all the states of model (features that are transferred over many layers of pre-trained models) and maps all the numerical representation into a label, however the Sequence Classifier utilize a slice of information from all the states and maps the representations into a label.

In summary, our contributions are:

\begin{itemize}
  \item Investigate the use two different classifier from different in nature in text classification engines on a standard dataset which allows us to compare results.
  \item Explore the advantages and disadvantages of each method.
  \item To label long texts which are longer than 256 token using token classification method.
\end{itemize}

\section{Data Description}

Modern natural language processing systems learn effective models with the help of annotated data as supervision. These models can only be utilized within that language because they are typically trained on data that is only available in that language (typically English). Since it is impractical to collect data in every language, cross lingual language understanding (XLU) and low-resource cross-language transfer have gained popularity. The Cross-lingual Natural Language Inference (XNLI) corpus is the extension of the Multi-Genre NLI (MultiNLI). There are 122k train, 2490 validation, and 5010 test samples in the dataset \cite{conneau2018xnli} in all 15 languages. The model is trained on the training set and the test set is used to find the best hyper-parameter and model. The validation set is used is an unseen set which is used later for testing the performance of the classifiers.

\begin{table}[!ht]
    \centering
    \caption{XNLI Data Stats}
    \begin{tabular}{|l|l|l|l|l|l|l|l|}
    \hline
        Set & Samples & Tokens & Token > 256 & Average Label & \% Label 0 & \% Label 1 & \% Label 2 \\ \hline
        Train & 392702 & 392702 & 66 & 1 & 33.33 & 33.33 & 33.33 \\ \hline
        Test  & 5010 & 5010 & 0 & 1 & 33.33 & 33.33 & 33.33 \\ \hline
        Validation & 2490 & 2490 & 0 & 1 & 33.33 & 33.33 & 33.33 \\ \hline
    \end{tabular}
        \label{XNLI Data Stats}
\end{table}

\begin{table}[!ht]
    \centering
    \caption{Sample data}
    \begin{tabular}{|p{0.45\linewidth} p{0.45\linewidth} p{0.05\linewidth} |}
    \hline
        \textbf{Premise} & \textbf{Hypothesis} & \textbf{Label} \\ \hline  \hline
        Conceptually cream skimming has two basic dimensions - product and geography . & Product and geography are what make cream skimming work . & 1 \\ \hline
        'you know during the season and i guess at at your level uh you lose them to the next level if if they decide to recall the the parent team the Braves decide to call to recall a guy from triple A then a double A guy goes up to replace him and a single A guy goes up to replace him' & You lose the things to the following level if the people recall . & 0 \\ \hline
        Gays and lesbians . & Heterosexuals . & 2 \\ \hline
    \end{tabular}
    \label{Sample data}
\end{table}

Table \ref{XNLI Data Stats} shows basic statistics about XNLI dataset on just English data such as the number of samples in each set, number overall tokens in all samples, number of texts whose number of tokens is greater than 256 and finally the distribution of all labels. XNLI dataset is used as sample set of data for testing the hypothesis of scoring regime on two different methods. The dataset is chosen since the label distribution and mean of labels are perfectly balanced.

Table \ref{Sample data} shows a sample set of text data which we used for classification purpose. Premise and Hypothesis text are combined and construct our sample text. The labels are integer encoded numbers which zero represents entailment, one represents neutral and 2 represents contradiction. This sample would be suitable for comparison study about the different nature of transformer classifiers.


\section{Background for Text Classification Engines}

The development of neural network-based natural language processing has had a significant impact on the current possibilities for text classification engines. A few preliminary corresponding comments on the meaning of some of the terms used as well as some additional methodological background will be provided beforehand because the terminology used in a machine learning context is frequently different from that used in psychometrics and is also occasionally ambiguous. Following that, more background information is provided along with two significant NLP techniques based on neural networks and their accompanying AES successes.

\subsection{Terms}
\begin{itemize}
  \item \textbf{Model}: A particular architecture for a machine learning which has a lot parameters needs to be tuned to find a mapping between a text and label. This can be pre-trained with a specific training dataset
  \item \textbf{Token}/Tokenization: A text sequence is an input to the model in NLP task. Making this text sequence into a chunk of words or sub words and map them to a series of numbers called tokenization.
  \item \textbf{Padding}: The only input sequences that neural networks accept are fixed size. The nature of text are variable in length to overcome the issue we add pad word in to a sequence which is less than maximum length size.
  \item \textbf{Embeddings}: The embeddings give the tokens a meaning by converting a single token into a high-dimensional vector.
  \item \textbf{Feature} : In text classification  regime, the features are referred to embedding of each token of a text sequence.
  \item \textbf{Language Model}: A language model is a model that predicts the next word, subword, or character in a text.
\end{itemize}

\subsection{Text Classification Modeling Approaches}
\begin{itemize}
  \item \textbf{Traditional Approaches} :  Bag of words approach (BOW) was commonly used in text classification for a number of years and the aim in these models are to extract features from the text and label the text based on those features \cite{chen2010unsupervised,leacock2003c,ramalingam2018automated}.
  \item \textbf{Recurrent Neural Nets (RNN)}: The gated recurrent units (GRU) \cite{cho2014learning}and the long short-term memory (LSTM) \cite{hochreiter1997long} are the two most common types of RNNs models. They used very often in the field of AES.
  \item \textbf{Transformer Models}: Transformer models are neural networks based on an attention mechanism \cite{vaswani2017attention}that are pre-trained on the large text data for the purpose of language modeling.
\end{itemize}

A transformer-based language models from the various architecture types (BERT, ELECTRA, ROBERTA, etc.) first undergo self-supervised training on sizable text corpora. They are then fine-tuned to a particular task (like classification, question/answering, token classification, generation, summarization) in a subsequent stage utilizing supervised learning and hand-coded labels. In the next sections we explain our methodology to use different transformer classifier on long or short texts.

\section{Long Text Tokenization}

Powerful neural network architectures known as Transformers \cite{vaswani2017attention, velivckovic2017graph, vig2019multiscale} have achieved state of art status in a number of machine learning domains, including Natural Language Processing (NLP), Neural Machine Translation (NMT) \cite{chen2018best} and text generation summarization. Unfortunately, a conventional Transformer scales quadratically with the number of tokens, making it unaffordable for a sequence larger than 512 tokens. For this problem, several solutions have been proposed in \cite{beltagy2020longformer, child2019generating, bello2019attention,gulati2020conformer}. Most methods limit the attention mechanism to focus on nearby neighborhoods or include structural priors on attention such as sparsity, pooling-based compression, clustering, or convolution techniques.

A dataset must first be preprocessed into an input format before it can be used to train a model. Your data must be transformed and put together into batches of texts. This preprocessing method is called tokenization, we use a tokenizer to break text into a series of tokens, then turn those tokens into numbers and put them together into certain ID numbers. Transformer architectures are trained on the large set of data with a restriction on the size of tokenized input. Most of transformers have a limit of 512 input tokens.

In this work we are not changing the model architecture to address the issues of long texts by increasing the limit size, however we are proposing to use the 256 limit size architecture using a method using overlapping segments. In the Token Classier model, we are introducing an stride hyper parameter that is the amount the text which we are overlapping. A single long text will be chunked and as $N$ number of texts of 256 token size, in such way that we track the overflow tokens.

\begin{figure}[hbt!]
\includegraphics[width=5.5in, height=1.2in ]{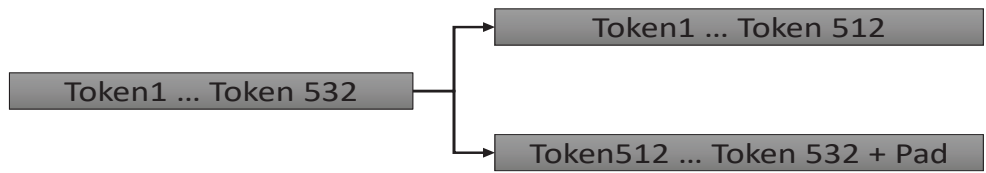}
\caption{Long Text Chuncking} \label{LongChuncking}
\end{figure}

In Figure \ref{LongChuncking}, the text containing 532 tokens, was passed through a tokenizer that returned tokenized inputs of length 512. Since the number of tokens in the text was more than the max length, the text was truncated to length 512, and the remaining tokens were put in the next entry. The last 20 token ids of the first entry were also present at the beginning of the second entry due to the use of the stride parameter. Since the second entry falls short of the max length of 512, padding was added to the second entry.

\section{Methodology}

\subsection{Token Classifier For Text Labeling}

\begin{figure}[hbt!]
\includegraphics[width=5.5in, height=2.2in ]{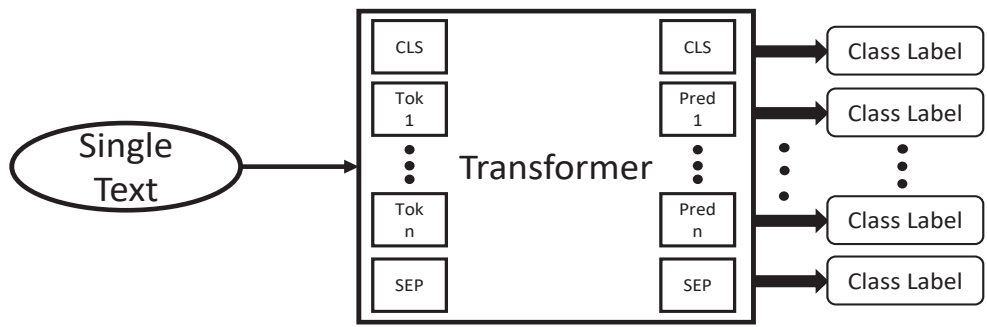}
\caption{Token Classifier Model} \label{TokenClassifier}
\end{figure}

The token classification model architecture is unquestionably the most popular one among all the Machine Learning techniques used in NLP. There has been a lot of work been done in the field of token classification like the the following research works;  dependency parsing \cite{milidiu2009token}, detecting label errors in Token Classification data \cite{wang2022detecting} and token sequence labeling vs. clause classification for English emotion stimulus detection \cite{oberlander2020token}. Tokenization, or simply identifying the beginning and end of the words and punctuation in the text, is the initial stage in token categorization. The next step is to decide which class (tag or token) belongs to. The choice of classes for issues like part-of-speech tagging is simple: they are the part-of-speech classes that each word may have, such as noun, verb, adjective, etc. Since one or more nearby tokens may appear to be a member of the same sort of phrasal chunk, but be a part of a different chunk. Other tasks, like chunking, necessitate subtle alterations when defining their classes. This would be useful for a name entity recognition system,

However, we propose to use the Token Classifier as a text classier that could label texts. The XNLI dataset is used in this case as a sample set that represents texts. Every single text has a label, that label is going to be assigned for every single token presented in the text shown in Figure \ref{TokenClassifier}. The model predicts label for every single token and finally the average of all the token labels will be assigned a final label.

For texts that the length is more than 256 tokens, the overflow option of the tokenizer is used and mapped all the overflow token to a unique id. Now we can train an text with more than 256 token by introducing the stride hyper parameter. Therefore, we could scan a long texts by starting the first 256 token and then use stride of $n$ to overlap the texts again and get another 256 chunks till the and long texts will be a chunk of 256 texts.

Finally, we can train the Token Classifier concept in token classification theme with the chunking mechanism which addresses the issue of using the first 256 tokens.

\subsection{Sequence Classifier For Text Labeling}

Text classification is a well-known and extensively studied job in Natural Language Processing (NLP) and text mining. It refers to the act of assigning one or more labels from a set of classes to a document \cite{sebastiani2002machine}. Simple binary classification, multi-class classification, and multi-label classification are examples of text classification problem types. Simple binary classification, for instance, asks if a document has positive sentiment or not. The latter allows for the assignment of many labels to a single document. Text Classification engines are the subcategory of text classification.

Tokenizing the text input is the initial stage in the classification process using a transformer model, just like in the other traditional methods. The chosen pretrained transformer has a strict binding effect on the tokenization's format. The next step is to use the Sequence Classifier model that will be fine-tuned on the set of labels.

\begin{figure}[hbt!]
\includegraphics[width=5.5in, height=2.2in ]{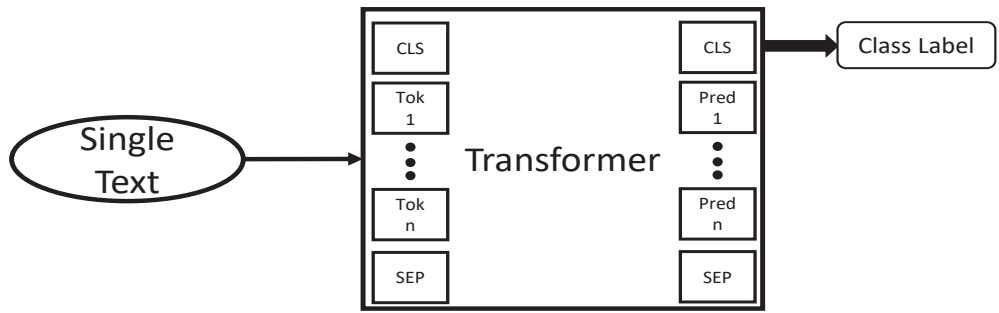}
\caption{Sequence Classifier Model} \label{SequenceClassifier}
\end{figure}

As you can see in Figure \ref{SequenceClassifier}, the first predicted token is used to map the features to the target labels however in the token classier all the prediction token are used in prediction the the final label for the text.

\section{Results}
The results of scoring using Sequence Classification and Token Classification are compared. First the pre-trained Electra model \cite{clark2020electra} is fine-tuned on training set. A grid search technique is used to find a best model on our metric. QWK metric is our main metric to evaluate the performance of the model. The hyper parameter that we are optimizing our models are:

\begin{itemize}
  \item Learning Rate
  \item Batch Size
  \item Stride
\end{itemize}

\subsection{Grid Search}

\begin{table}[!ht]
    \centering
    \caption{Grid Search - Sequence Classification}
    \begin{tabular}{|l|l|l|l|l|l|l|}
    \hline
        experiment\_id & epoch & num label & test acc & test QWK & batch size & lr  \\ \hline
        3 & 1 & 3 & 0.805 & 0.766 & 60 & 0.000035 \\ \hline
        2 & 1 & 3 & 0.807 & 0.766 & 50 & 0.000025 \\ \hline
        1 & 1 & 3 & 0.805 & 0.766 & 50 & 0.000035 \\ \hline
        4 & 2 & 3 & 0.799 & 0.760 & 60 & 0.000025 \\ \hline
    \end{tabular}
    \label{Grid Search Sequence Classification}
\end{table}

\begin{table}[!ht]
    \centering
    \caption{Grid Search - Token Classification}
    \begin{tabular}{|l|l|l|l|l|l|l|l|l|}
    \hline
        experiment & epoch & num label & test acc & test QWK & stride & max token & batch size & lr \\ \hline
        8 & 4 & 3 & 0.825 & 0.800& 250 & 256 & 12 & 0.000035 \\ \hline
        4 & 5 & 3 & 0.810 & 0.792 & 200 & 256 & 10 & 0.000025 \\ \hline
        5 & 3 & 3 & 0.821& 0.790 & 250 & 256 & 10 & 0.000025 \\ \hline
        1 & 4 & 3 & 0.819 & 0.789 & 200 & 256 & 10 & 0.000035 \\ \hline
        3 & 5 & 3 & 0.815 & 0.789 & 300 & 256 & 10 & 0.000035 \\ \hline
        2 & 7 & 3 & 0.817 & 0.788 & 250 & 256 & 10 & 0.000035 \\ \hline
        6 & 10 & 3 & 0.817 & 0.787 & 300 & 256 & 10 & 0.000025 \\ \hline
        7 & 4 & 3 & 0.814 & 0.780 & 200 & 256 & 12 & 0.000035 \\ \hline
    \end{tabular}
    \label{Grid Search Token Classification}
\end{table}

Table \ref{Grid Search Sequence Classification} shows the result of training a Sequence Classier Electra model on the training set and save the best model on the test set. The highest test QWK metric on the Sequence Classification method 0.76.

Table \ref{Grid Search Token Classification} shows the result of training a Token Classier Electra model on the train set and save the best model on the test set. As you can in the token classification, stride is another hyper parameter that is introduced to deal with a long text. The highest test QWK metric on the the token classification method 0.80.

Now it is the time to evaluate the best models on the validation set which the network has not seen before to draw a final decision on the performance of each model.

\subsection{Validation Results}

Final models are tested on the validation set and Table \ref{Validation Results} shows each model performance on the validation set. As you can see the Token Classifier has higher accuracy (test acc and val acc) and perform well on scoring texts. Beside QWK, standardized mean difference (SMD) is used to measure efficacy in terms of a continuous measurement (lower better).

\begin{table}[!ht]
    \centering
    \caption{Validation Results}
    \begin{tabular}{|l|l|l|l|}
    \hline
        Model & val acc & val qwk & val smd \\ \hline
        Token Classifier & 0.827 & 0.803 & 0.043 \\ \hline
        Sequence Classifier & 0.814 & 0.781 & 0.052 \\ \hline
    \end{tabular}
    \label{Validation Results}
\end{table}

\section{Conclusion and Future Work}

In this work, the performance between the Sequence Classifiers and Token Classifiers are studied. The token classification model has better performance on text classification and labeling texts. The use of all of the prediction sequences that models’ outputs help to extract more information for predicting labels. The sequence classifier is the most common used model for scoring  but its performance is not as well as the token classification method, because sequence classifier just uses a first sequence prediction token which ignores all the other sequences information and then maps the texts into its label. Certainly, the use all the sequences that are extracted from the features provided better performance because the base model are exactly the same for the this study . For future work, different type of aggregation in token classification can be explored instead of using a mean of label for every single token. Also, we could verify the performance on the Kaggle essay data set.

\bibliographystyle{IEEEtran}
\balance
\bibliography{mybib_Token.bib}

\end{document}